\documentclass[runningheads]{llncs}
\usepackage{graphicx}

\usepackage{tikz}
\usepackage{comment}
\usepackage{amsmath,amssymb} 
\usepackage{color}
\usepackage{booktabs}

\usepackage[accsupp]{axessibility}  

\usepackage[colorlinks=true,linkcolor=black,citecolor=green,urlcolor=blue,]{hyperref}
\usepackage[doipre={doi:~}]{uri}
\usepackage[numbers,sort&compress]{natbib}

\usepackage{indentfirst}
\usepackage{bbding} 
\begin{document}
\sloppy
\pagestyle{headings}
\mainmatter
\def\ECCVSubNumber{100}  

\title{ReIDTrack:Multi-Object Track and Segmentation Without Motion}
\titlerunning{CVPR-23 submission ID \ECCVSubNumber} 
\authorrunning{CVPR-23 submission ID \ECCVSubNumber} 
\author{Kaer Huang\textsuperscript{1}, Bingchuan Sun\textsuperscript{1}, Feng Chen\textsuperscript{1}, Tao Zhang\textsuperscript{2}, \\ Jun Xie\textsuperscript{1}, Jian Li\textsuperscript{1}, Christopher Walter Twombly\textsuperscript{1}, Zhepeng Wang\textsuperscript{1}\textsuperscript{\Envelope}
}
\institute{\textsuperscript{1}Lenovo   \textsuperscript{2}Tsinghua University  }

\maketitle

\begin{abstract}

In recent years, dominant Multi-object tracking (MOT) and segmentation (MOTS) methods mainly follow the tracking-by-detection paradigm. Transformer-based end-to-end (E2E) solutions bring some ideas to MOT and MOTS, but they cannot achieve a new state-of-the-art (SOTA) performance in major MOT and MOTS benchmarks. Detection and association are two main modules of the tracking-by-detection paradigm. Association techniques mainly depend on the combination of motion and appearance information. As deep learning has been recently developed, the performance of the detection and appearance model is rapidly improved. These trends made us consider whether we can achieve SOTA based on only high-performance detection and appearance model. Our paper mainly focuses on exploring this direction based on CBNetV2 with Swin-B as a detection model and MoCo-v2 as a self-supervised appearance model. Motion information and IoU mapping were removed during the association. Our method wins 1st place on the MOTS track and wins 2nd on the MOT track in the CVPR2023 WAD workshop. We hope our simple and effective method can give some insights to the MOT and MOTS research community. Source code will be released under this git repository \url{https://github.com/CarlHuangNuc}.

\end{abstract}

\section{Introduction}

Object tracking is one of the fundamental tasks in computer vision, which
used to build instance-level correspondence between frames and output trajectories with boxes or masks \cite{yan2022towards}. MOT and MOTS tasks aim to simultaneously process detecting, segmenting, and tracking object instances in a given video \cite{wu2022defense2}. It can be used in video surveillance, autonomous driving, video understanding, etc. 
\par
Current mainstream methods follow the tracking-by-detection paradigm\\ \cite{liang2022rethinking,lu2020retinatrack,sun2020transtrack,wu2021track}. Until recent years, Transformer-based E2E solutions brought new ideas to MOT and MOTS research areas \cite{dosovitskiy2020image,zeng2021motr,cheng2021per,cheng2022masked}, but their performance could not reach SOTA in major MOT and MOTS benchmarks. Detection and association are two main modules of the tracking-by-detection paradigm. Association techniques mainly depend on the combination of motion and appearance information \cite{zhang2021fairmot,pang2021quasi}. As deep learning developed, appearance and detection models get rapid improvement in performance. At the same time, the difficulty of the autonomous vehicle dataset includes low video frame rate, fast movement, and large displacement. The traditional association methods based on IoU and motion do not perform well in this kind of situation. 
\par
The challenge of association based on motion information, made us consider whether we can archive SOTA only based on high-performance detection and appearance model. Our paper tried to explore this direction. We use CBNetV2 Swin-B \cite{liang2021cbnetv2} as the detection model and self-supervised learning MoCo-v2 \cite{he2020momentum} as a high-quality appearance model. We removed all motion information, including the Kalman filter and IoU mapping, and archived SOTA on BDD100K dataset. Our method wins 1st Place in the CVPR2022 WAD BDD100K MOT challenge, and 1st Place in the ECCV2022 SSLAD track 4 BDD100K challenges, including MOT, MOTS, SSMOT, and SSMOTS tracks. Our method also wins 1st Place in the CVPR2023 WAD BDD100K MOTS challenge. We hope our simple and effective method can give some insight into the MOT and MOTS research community.


\section{Related Work}
\textbf{Multi Object Tracking (MOT)} is a very general algorithm and has been studied for many years. The mainstream methods follow the tracking-by-detection paradigm \cite{liang2022rethinking,lu2020retinatrack,sun2020transtrack,wu2021track}. With the development of deep learning in recent years, the performance of the detection model is improved rapidly. Currently, most of the work relies on YOLOX \cite{yan2022towards,zhang2021bytetrack}. Our method selected a stronger performance network CBNetV2 \cite{liang2021cbnetv2}  which is used to verify the potential of the detector in our hypothesis. Another important component of MOT is an association strategy. 
Popular association methods include motion-based (IoU matching, Kalman filter) \cite{bewley2016simple}, appearance-based (ReID embedding) \cite{wang2021different}, transformer-based \cite{zeng2021motr}, or the combination of them \cite{zhang2021fairmot,pang2021quasi}. Our methods remove all motion information and use only a high-performance appearance model.
\par

\textbf{Multi Object Tracking and Segmentation (MOTS)} is highly related to MOT by changing the form of boxes to fine-grained mask representation  \cite{yan2022towards}. Many MOTS methods are developed upon MOT trackers \cite{ke2021prototypical,voigtlaender2019mots,huang2023multi}. Our ideas are similar to theirs. A mask header was added on the basis of the MOT network in  our MOTS solution. 

\par

\textbf{Self-Supervised Learning} has made significant progress in representation learning in recent years. Contrastive learning, one of the self-supervised learning methods such as MoCo\cite{he2020momentum}, SimCLR\cite{chen2020simple},  BYOL\cite{grill2020bootstrap},  etc, has performance that is getting closer to results of supervised learning methods in ImageNet dataset. We leveraged Momentum Contrastive Learning (MoCo-v2)\cite{he2020momentum} to train a new appearance embedding model without using tracking annotations. The technique not only meets the requirements of SSMOT and SSMOTS but also improves the performance of the appearance model.

\section{Method}
\begin{figure}
    \centering
    \includegraphics[width=\textwidth]{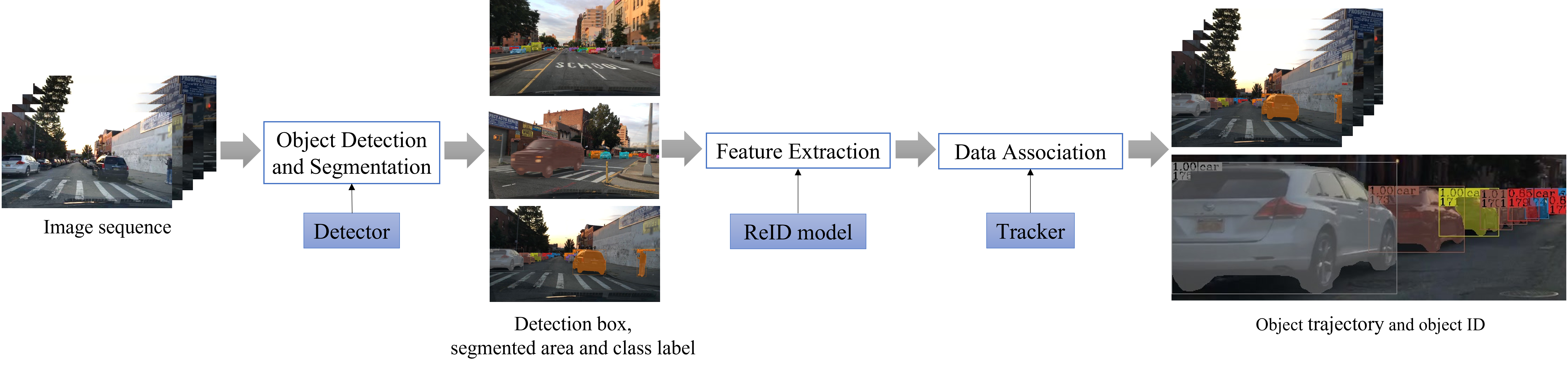}
    \caption{Our framework}
    \label{fig:overview}
\end{figure}

The overview of our framework is shown in figure \ref{fig:overview}. The framework is based on a tracking-by-detection paradigm. Object bounding boxes are detected in each image by a detector in MOT. In MOTS, a segmentation head is added to the detector to extract binary masks within each detected box. A ReID model extracts features from the bounding boxes. Then, a tracker process the data association to match the object ID in the image sequence. 

\subsection{Detection and Segmentation}
We applied CBNetV2 architecture to connect two Swin-B with FPN backbones in parallel. Features from high and low levels of the backbones are integrated to improve detector performance. The HTC detection head was used to predict box and binary mask. The mask head is trained with a multi-step training strategy. Firstly, the model was trained for box detection by using a relatively large number of box-labeled data. Then, the whole network with a mask branch was fine-tuned based on MOTS labeled dataset. In addition, a multi-class NMS threshold is applied to reduce the data imbalance problem.

\subsection{Re-Identification}
We used Unitrack as a ReID module for MOT and MOTS. Our appearance model for this framework is MoCo-v2 with ResNet50 backbone. The model extracts feature representations from detected boxes. The tracklet features are weighted by the detection score and combined within $\tau$ frames to maintain the object representation during occlusion. The weighted feature $\hat{e_j}$ combined tracklet feature $e_j$ which is weighted by the detection score $s_j$ from the previous $\tau$ frames. 
\par
\begin{equation}
    \hat{e_j}=\frac{\sum^\tau_{t=1}e^t_j\times s^t_j}{\sum^\tau_{t=1}s^t_j}
\end{equation}
\par
$\hat{e_j}$ is further used for computing ReID distance in the data association.

\subsection{Tracking}
ByteTrack method, which divides detection boxes into high and low detection scores for data association, is used in our framework. Firstly, the high score boxes are used to associate with the tracklet. The remained high-score boxes will be kept as tentative boxes, which will become a new tracklet after appearing for 2 consecutive frames. Then, the low score boxes are used to find the match with the remained tracklet. From our experiments, using ReID distance has the best results in all high and low-score box associations. Then, the Hungarian algorithm uses the distance to assign the tracking ID in each association step. For the lost and occluded tracklets, they are kept within 10 frames.

\section{Experiments}
 In this section, we introduce the dataset and evaluation metrics. Then, we explain our implementation details for experiments. Finally, we report the main results on the CVPR2023 BDD100K Challenges test server and ablation study of major methods.  
 \subsection{Dataset and Evaluation Metrics}
  We conducted experiments on the BDD100K dataset which is a large-scale autonomous driving video dataset with 100K driving videos (40 seconds each). BDD100K provides multi-task annotations for MOT and MOTS. The MOT dataset contains 1400 and 200 videos with annotation for training and validation, respectively, and 400 videos for testing. The MOTS dataset contains 154 and 32 videos with annotation for training and validation, respectively, and 37 videos for testing.

 Mean Track Every Thing Accuracy (TETA, mean of TETA of the 8 categories) as our primary evaluation metric for ranking. These competitions also employ mean Higher Order Tracking Accuracy (HOTA) order, mean Multiple Object Tracking Accuracy (mMOTA) and mean ID F1 score (mIDF1), which are previously used as the main metrics. For MOTS, we use the same metrics set as MOT. The only difference lies in the computation of distance matrices. In MOT, it is computed using box IoU, while for MOTS the mask IoU is used.

 \subsection{Implementation Details}
\textbf{Detector}. CBNetV2 was trained on both BDD100K object detection and MOT dataset. The Swin-B backbone was initiated by a model pre-trained on ImageNet-22K. We applied multi-scale augmentation to scale the shortest side of images to between 640 and 1280 pixels and applied random flip augmentation during training. The optimizer is AdamW with an initial learning rate of 1e-6 and weight decay of 0.05. We trained the model on 4 A100 GPUs with 1 image per GPU for 10 epochs. At inference time, we resize the image size to 2880x1920 to better detect the small objects. We applied the multi-class NMS thresholds 0.6, 0.1, 0.5, 0.4, 0.01, 0.01, 0.01, and 0.4 for pedestrian, rider, car, truck, bus, train, motorcycle, and bicycle class, respectively.

\indent\textbf{Segmentation Head}. The backbone, neck, and detection head was initiated by MOT detector. Then, we fine-tuned the MOTS detector with BDD100K instance segmentation and MOTS dataset. The AdamW optimizer was set the initial learning rate of 5e-7 and weight decay of 0.05. We trained the model on 4 A100 GPUs with 1 image per GPU for 20 epochs. 

\indent\textbf{ReID}. The backbone of ReID is pre-trained on ImageNet-1K. Then, we fine-tuned the backbone by using MoCo-v2 on BDD100K dataset. The training dataset contains cropped object images according to bounding box labels from MOT dataset. The optimizer is SGD with weight decay of 1e-4, momentum factor of 0.9, and initial learning rate of 0.12. We trained the model on 4 A100 GPUs with 256 images per GPU.

\indent We do not rely on the tracking annotations when training the detector, segmentation head, and ReID model, thus our method can be applied to SSMOT and SSMOTS.

\indent\textbf{Tracker}. Our method is generally similar to ByteTrack, but we used ReID to match high and low detection boxes.  We set the high detection score threshold to 0.84 and the low detection score threshold to 0.3.

 \subsection{Main Results}
We evaluated the performance of our method on BDD100K MOT and MOTS test set. We achieve 57.07 and 54.09 mHOTA in BDD100K MOT and MOTS, as shown in Table \ref{BDD100K MOT test set result} and \ref{BDD100K MOTS test set results}. 
\begin{table}[!ht]
    \centering
    \caption{Comparison with other methods on \textbf{BDD100K MOT} test set. \textbf{Bold} represents the best metrics.}
    \label{BDD100K MOT test set result}
    \begin{tabular}{c c c c c c c c}
    \hline
        Team & mTETA & mHOTA & mMOTA & mIDF1 & mDetA & mAssA & mMOTP \\ \hline
        vdig & 58.46 & 46.3 & 38.1 & 55.2 & 41.0 & 53.9 & 81.1 \\
        \textbf{Ours} & \textbf{57.07} & \textbf{49.2} & \textbf{43.0} & \textbf{59.5} & \textbf{43.9} & \textbf{56.4} & \textbf{81.4} \\ 

        CMSQ & 54.82 & 44.67 & 40.11 & 53.52 & 39.28 & 52.33 & 82.58 \\
        HELLORPG & 53.62 & 41.99 & 34.10 & 49.53 & 32.16 & 56.52 & 80.53 \\
        MTIOT & 53.31 & 42.51 & 35.79 & 50.49 & 32.89 & 56.70 & 80.65 \\
        hua & 49.83 & 41.62 & 36.47 & 51.67 & 35.79 & 50.56 & 76.72 \\
        LittleBoss & 42.47 & 29.29 & 23.69 & 34.72 & 22.69 & 39.92 & 76.25 \\
         
       \hline
    \end{tabular}
\end{table}

\begin{table}[!ht]
    \centering
    \caption{Comparison with other methods on \textbf{BDD100K MOTS} test set.}
    \label{BDD100K MOTS test set results}
    \begin{tabular}{c c c c c c c c}
    \hline
        Team & mTETA & mHOTA & mMOTA & mIDF1 & mDetA & mAssA & mMOTP \\ \hline
        \textbf{Ours} & \textbf{54.09} & \textbf{44.0} & \textbf{41.1} & \textbf{54.9} & \textbf{39.3} & \textbf{50.8} & \textbf{69.7} \\ 
        vdig & 51.17 & 44.06 & 41.81 & 56.20 & 39.62 & 50.30 & 69.10 \\ 
        CMSQ & 47.93 & 39.70 & 31.62 & 49.37 & 34.47 & 46.78 & 68.94 \\
        hua  & 46.63 & 40.00 & 32.59 & 50.34 & 35.49 & 46.70 & 67.41 \\ 
        Host\_4901\_Team & 45.73 & 39.2 & 31.9 & 50.4 & 33.8 & 46.3 & 66.5 \\
        ksghsharma  & 44.96 & 38.67 & 34.02 & 50.05 & 33.14 & 45.97 & 66.47 \\ 
        
        \hline
    \end{tabular}
\end{table}

 \section{Conclusions}
 In this paper, we propose a simple yet effective tracking-by-detection framework for multi-object tracking (MOT) and segmentation (MOTS). We discard the motion information and only use the appearance embeddings to associate the objects. The training of detection and appearance models does not rely on tracking annotations which can be costly to obtain. Our method achieves first place in CVPR2023 WAD BDD100K MOTS Challenge and second place in CVPR2023 WAD BDD100K MOT Challenge.

\renewcommand{\bibname}{References}
\bibliographystyle{splncs04}
\bibliography{egbib}
\end{document}